\documentclass[runningheads]{llncs}
\usepackage{graphicx}
\usepackage{epsfig}

\usepackage{verbatim}
\usepackage{helvet}
\usepackage{courier}
\usepackage{amsfonts}
\usepackage{subfigure}
\usepackage{caption}
\graphicspath{ {./images/} }
\usepackage{booktabs}
\usepackage{threeparttable}
\usepackage{multicol}
\usepackage{multirow}
\usepackage{latexsym}
\usepackage{tabularx}
\newcolumntype{L}[1]{>{\raggedright\arraybackslash}p{#1}}
\newcolumntype{C}[1]{>{\centering\arraybackslash}p{#1}}
\newcolumntype{R}[1]{>{\raggedleft\arraybackslash}p{#1}}

\usepackage{tikz}
\usepackage{comment}
\usepackage{amsmath,amssymb} 
\usepackage{color}

\usepackage[accsupp]{axessibility}  

\usepackage[width=122mm,left=12mm,paperwidth=146mm,height=193mm,top=12mm,paperheight=217mm]{geometry}
\usepackage[pagebackref,breaklinks,colorlinks,bookmarks=false]{hyperref}

\usepackage[capitalize]{cleveref}
\crefname{section}{Sec.}{Secs.}
\Crefname{section}{Section}{Sections}
\Crefname{table}{Table}{Tables}
\crefname{table}{Tab.}{Tabs.}

\begin{document}
\pagestyle{headings}
\mainmatter
\def\ECCVSubNumber{2570}  

\title{Towards Self-Supervised Category-Level Object Pose and Size Estimation}

\titlerunning{SelfCatPS}
%
\author{Yisheng He\inst{1} \and Haoqiang Fan\inst{2} \and Haibin Huang\inst{3} \and Qifeng Chen\inst{1} \and Jian Sun\inst{2}}
\authorrunning{Y. He et al.}
%
\institute{HKUST \and Megvii Technology \and Kuaishou Technology}

\maketitle

\begin{abstract}


In this work,  we tackle the challenging problem of category-level object pose and size estimation from a single depth image. Although previous fully-supervised works have demonstrated promising performance, collecting ground-truth pose labels is generally time-consuming and labor-intensive. Instead, we propose a label-free method that learns to enforce the geometric consistency between category template mesh and observed object point cloud under a self-supervision manner. Specifically, our method consists of three key components: differentiable shape deformation,  registration, and rendering. In particular, shape deformation and registration are applied to the template mesh to eliminate the differences in shape, pose and scale. A differentiable renderer is then deployed to enforce geometric consistency between point clouds lifted from the rendered depth and the observed scene for self-supervision. We evaluate our approach on real-world datasets and find that our approach outperforms the simple traditional baseline by large margins while being competitive with some fully-supervised approaches.


\end{abstract}
\section{Introduction}

\newcommand{\cmpMDL}{1}

\begin{figure}
    \centering
    \includegraphics[scale=0.56]{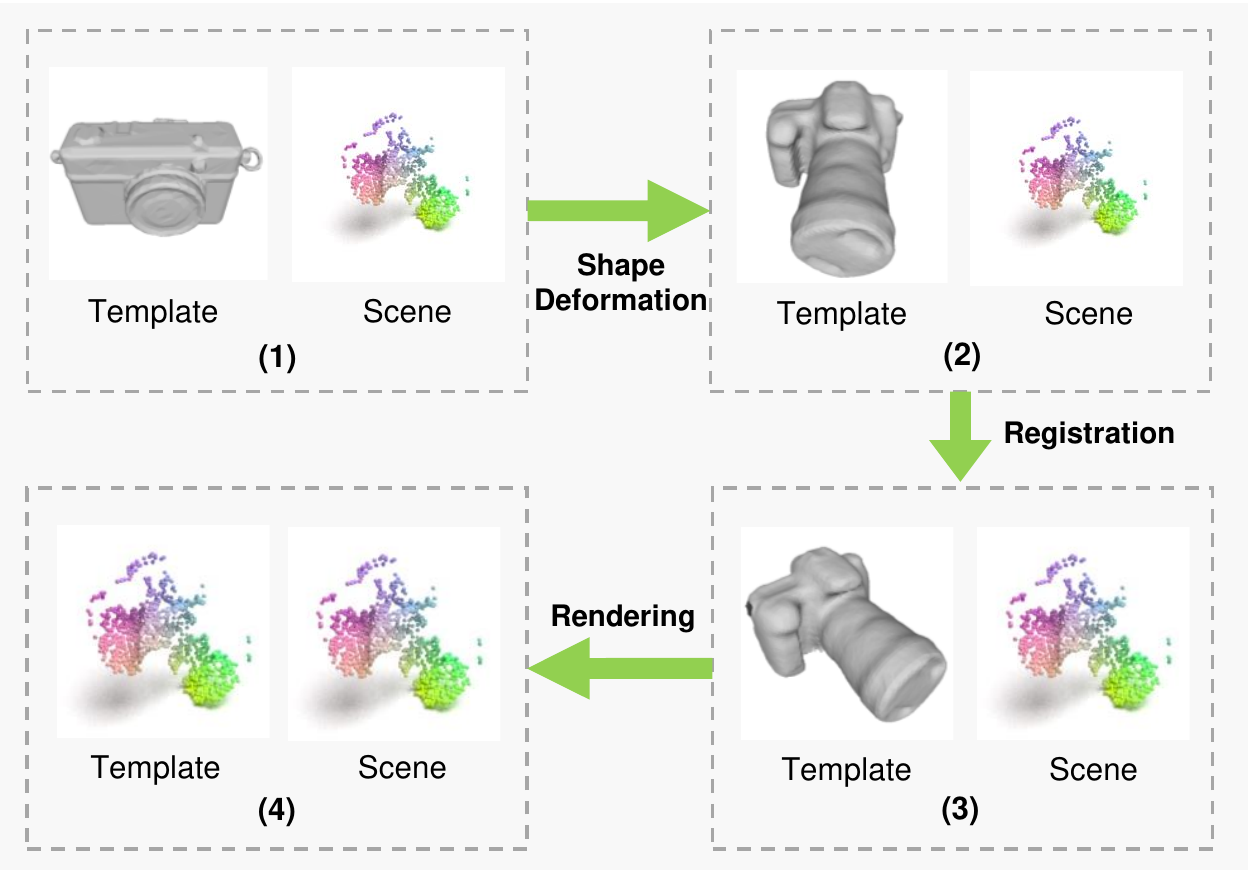}
    \caption{
        Challenges to self-supervised category-level pose and size estimation reside in the entangled shape, pose, scale, and partial-to-complete inconsistencies between the category template mesh and scene point cloud, shown in (1). We leverage differentiable shape deformation, registration, and rendering to disentangle them step by step and enforce geometric consistency between the rendered and observed scene point clouds for self-supervised learning.
    }
    \label{fig:intro}
\end{figure}

Category-level pose and size estimation is an important component in various real-world applications, including autonomous driving \cite{geiger2012we,chen2017multi,xu2018pointfusion}, augmented reality \cite{marchand2015pose}, and robotic manipulation \cite{collet2011moped,tremblay2018deep,he2020pvn3d}. The goal of this task is to predict the translation, orientation and size of novel objects in the same category.
Contrary to instance-level pose estimation \cite{he2020pvn3d,he2021ffb6d} for seen objects with CAD models, category-level algorithms are required to handle various objects within the same category and generalize to novel objects during inference. The intra-class variation of objects' shape and color makes this new problem more challenging. 

Recently, with the development of deep learning techniques, various data-driven approaches \cite{wang2019normalized,tian2020SPD,chen2021fsnet} have been introduced to tackle the problem and achieve significant improvement over traditional algorithms \cite{besl1992_ICP}. Since no instance-level object CAD models are provided to serve as the reference frame, two kinds of representation are introduced to define the category-level pose and size. One line of works \cite{wang2019normalized,chen2020CASS} utilize implicit Normalized Object Coordinate Space (NOCS) to map all possible object instances into the same representation. However, explicit shape variations of different objects in the same category are lost in such representation, which limits the performance of pose estimation. More recent works \cite{tian2020SPD,Chen_2021_SGPA}, instead, represent a category of objects by a normalized prior shape, which is then deformed to obtain the exact shape of novel objects for particular pose and 
size estimation. 

Despite compelling results being achieved, existing data-driven approaches heavily rely on large-scale datasets with ground-truth pose labels for fully-supervised learning. Unlike many 2D vision tasks, i.e., classification and segmentation, acquiring real-world 6D pose annotations is much more labor-intensive, and error-prone \cite{hodan2017tless,kaskman2019homebreweddb,wang2020self6d}, which hinders learning-based algorithms from scaling up. On the other hand, large-scale unlabeled depth data becomes easy to obtain with the rising prevalence of depth sensors, which provokes us to consider an interesting research question: How to leverage this data for self-supervised learning of category-level object pose and scale estimation? 
 
In this work, we propose a novel self-supervised framework for label-free pose and scale estimation. Given the category template mesh, our key idea is to leverage differentiable shape deformation, registration, and rendering to enforce the \textit{geometric consistency} between the rendered and the observed object point cloud for self-supervision. As shown in Figure \ref{fig:intro}, challenging obstacles to this idea resides in the entangled shape, pose, scale, and partial-to-complete inconsistency between the observed point cloud and the category template mesh. To path the way, we first introduce a pose- and scale-invariant shape deformation network to deform the template mesh to the target shape and eliminate the shape inconsistency. This module is supervised by the target object's complete coarse point cloud in the same pose and scale as the template mesh. Then, taking points sampled from the deformed mesh and the observed scene, a partial-to-complete registration network is applied for pose and scale estimation. Finally, with the predicted pose and scale parameters, we scale and transform the deformed mesh to the scene and leverage a differentiable renderer to render the visible parts and eliminate the partial-to-complete inconsistency. In this way, we can enforce the \textit{geometric consistency} between point clouds lifted from the rendered depth and the observed scene for self-supervision of pose and scale estimation.

We further conduct experiments on real-world datasets to validate the proposed approach. Experimental results show that our approach outperforms the simple traditional baseline significantly while competitive with some full-supervised approaches.

To summarize, the main contributions of this work are:
\begin{itemize}
\renewcommand{\labelitemi}{\textbullet}
\item A novel label-free learning framework towards self-supervised category-level object pose and scale estimation via differentiable shape deformation, registration, and rendering;
\item A novel pose- and scale-invariant mesh deformation network and a multi-scale geometry enhancement mechanism that fully leverages local and global geometric information for deformation guidance;
\item Extensive experiments demonstrate the efficacy of our approach against traditional and fully-supervised approaches;
\item In-depth analysis to understand various design choices of our system.
\end{itemize}
\section{Related Work}

\subsection{Fully-Supervised Category-Level Pose and Size Estimation}

NOCS-based approaches \cite{wang2019normalized,chen2020CASS} utilize normalized object coordinate space to represent objects within the same category for pose estimation. Shape-prior-based methods \cite{tian2020SPD,wang2021CRNet,Chen_2021_SGPA} jointly deform category-level prior point clouds and establish dense correspondence to recover pose parameters. Direct regression approaches \cite{chen2021fsnet,Lin_2021_dualposenet} are also introduced to the problem. However, these algorithms heavily rely on the labor-intensive ground-truth pose annotations for training.

\subsection{Self-Supervised Pose Estimation}
Given its practical importance, several works have looked into applying self-supervised techniques for \textit{instance-level} pose estimation. 
\cite{wang2020self6d,sundermeyer2020augmented} fine-tunes pre-trained model from synthesis data on real data through visual and geometric alignment. However, the reliance on precise instance-level object CAD model hinders the application of these frameworks to the \textit{category-level} problem. \cite{li2021leveraging} leverage SE(3) equivariance point cloud network for joint point shape reconstruction and 6D pose estimation. However, they ignore the object scale parameters and focus on rotation and translation estimation. While the object size information is useful in many real-world applications (i.e., robot manipulation), our work targets the more challenging pose and size estimation. Besides, before being applied to real-world data, existing works \cite{li2021leveraging,wang2020self6d,sundermeyer2020augmented} require large-scale synthesis data rendered from precise CAD models for network pre-training. While collecting precise object CAD models for new categories is labor-intensive, our framework is designed to be directly optimized on real-world data. To the best of our knowledge, we are the first to propose a self-supervised category-level \textit{pose} and \textit{size} estimation framework that can be directly optimized on unlabeled real depth images without any pre-training.

\subsection{Shape Deformation}
Shape deformation techniques are applied for object reconstruction from single images or point clouds. Learning-based algorithms can be divided into volumetric warps \cite{jack2018learning,kurenkov2018deformnet,yumer2016learning,krull2015learning}, cage deformations \cite{yifan2020neural}, vertex-offsets-based \cite{wang20193dn} and flow-based approaches \cite{jiang2020shapeflow}. Most related to our work, 3DN\cite{wang20193dn} regresses vertex offsets to deform mesh from global features extracted from the source meshes and target point clouds. However, it assumes small pose and scale differences and are sensitive to such variations. Similar to ours, FS-Net \cite{chen2021fsnet} utilizes shift and scale-invariant 3DGCN to extract global features for oriented point cloud reconstruction. However, the reconstructed points are sparse without topology information, causing artifacts in rendered depths. The orientation of points is not disentangled as well. In contrast, we propose a pose- and scale-invariant framework to extract features for mesh deformation. Moreover, to fully leverage the local and global geometric information from the observed partial point cloud, we introduce a novel multi-scale cross attention module based on transformers to enhance the shape deformation.

\section{Method}
Given the partial point cloud of a novel object from the scene and its category template mesh, the target is to obtain its pose and scale parameters. Though the observed object point cloud and the template mesh are within the same category, they are inconsistent in three dimensions, making the problem challenging: 1. They are in different shapes. 2. Their poses and sizes are different.
3. The observed point cloud is partial while the prior mesh model is complete.
To design a self-supervised framework for pose and size estimation, we need to eliminate these inconsistencies to establish consistent constrain for network supervision.

\begin{figure}
  \centering
  \hspace{-2mm}\includegraphics[width=1.02\linewidth]{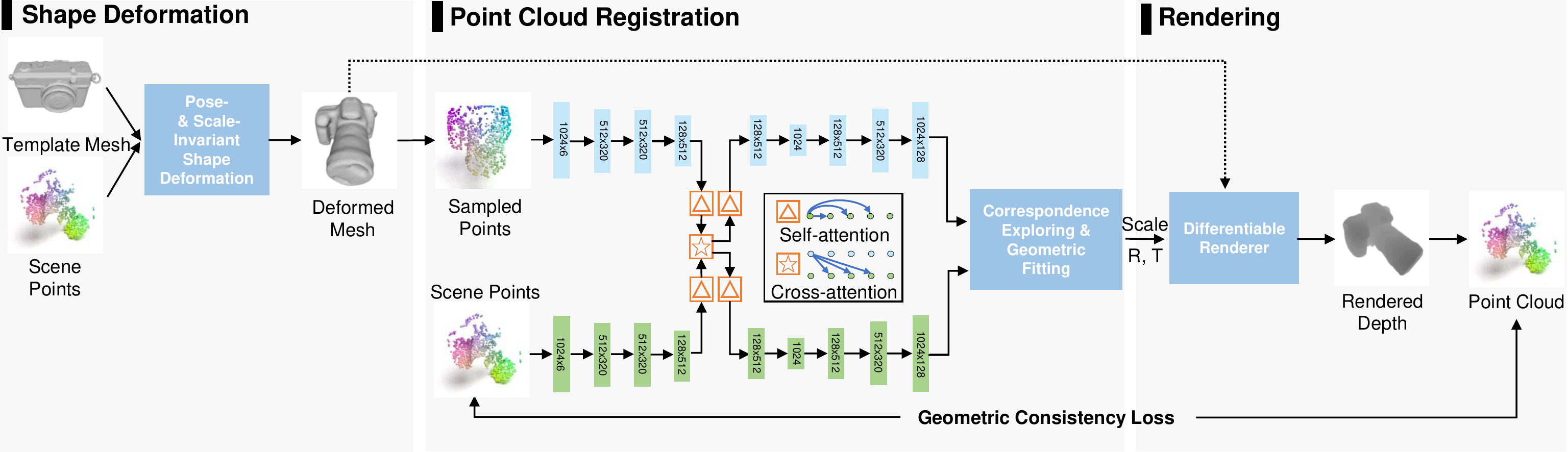}
  \caption{
\textbf{The overall pipeline of our framework.} Our pipeline consists of three main steps. Firstly, a pose- and scale-invariant shape deformation network is utilized to deform the template mesh to the shape of the observed object. Then, a point cloud registration network is applied to explore correspondence and estimate the pose and scale parameters. Finally, the deformed mesh is rendered by a differentiable renderer, and the geometric consistency between the rendered and the observed scene point cloud is established for self-supervised learning.
  }
  \label{fig:network_full}
\end{figure}

\subsection{Overview}
As shown in Figure \ref{fig:network_full}, we solve the problem in three steps, by differentiable mesh deformation, point cloud registration, and mesh rendering. Each step tackles one of the above inconsistency problems and eventually establishes the geometric consistent constrain between the rendered depth and the observed scene point cloud for self-supervision. Specifically, in the first step, a pose- and scale-invariant shape deformation network is utilized to deform the prior mesh model to the target object s.t. the observed partial point cloud to eliminate the shape inconsistency. Inspired by point cloud registration pipelines \cite{huang2021predator,el2021unsupervisedr}, in our second step, we estimate pose and size parameters by extracting feature descriptors, finding correspondences, and estimating the best alignment. Finally, we transform the deformed mesh by the predicted scale and pose to the camera coordinate system and utilize differentiable rendering to eliminate the partial-to-complete inconsistency. The rendered depth map is lifted back to a partial point cloud that should be geometric consistent with the observed scene point cloud, which can serve as the supervision signal to our framework. Each module in our pipeline is differentiable by design, and the whole framework can be optimized end-to-end.

\begin{figure}
  \centering
  \includegraphics[width=1.02\textwidth]{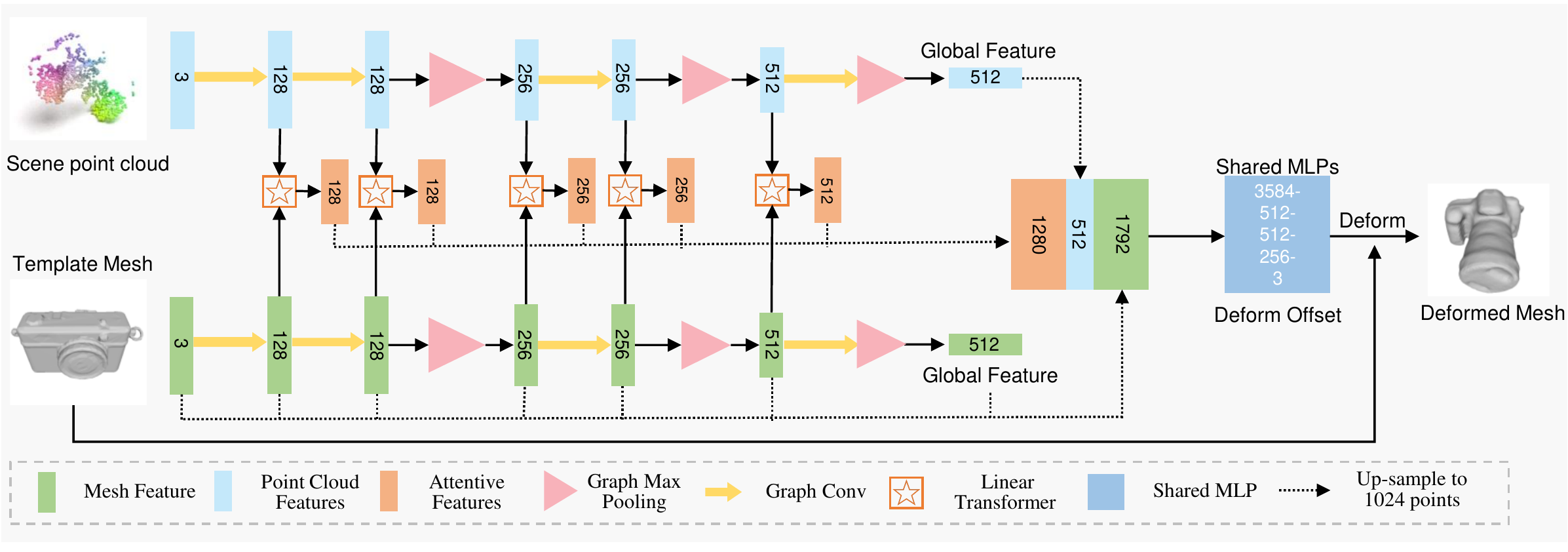}
  \caption{
\textbf{The pose- and scale-invariant shape deformation network}. We apply shift- and scale-invariant 3DGCN \cite{lin2020_3DGC} for feature extraction. The template mesh's local and global geometric features and the pose- and scale-invariant point cloud global feature are directly applied for shape deformation. To fully leverage different levels of local geometric information residing in the partial object point cloud to guide the shape deformation, a novel multi-scale geometry enhancement mechanism based on Linear Transformers \cite{wang2020linear_trans} is also utilized to extract more geometric details and guide better shape deformation.
  }
  \label{fig:network_deform}
\end{figure}

\subsection{Pose- and Scale-Invariant Shape Deformation}  
Given the observed partial point cloud, $P_{scn}$ of a novel object and its normalized category template mesh, $M_{tp}$, our task is to deform $M_{tp}$ to the target mesh $M_{tgt}$, where $M_{tgt}$ is with the same shape as $P_{scn}$ while kept in the same pose and scale as $M_{tp}$. We represent $P_{scn} \in \mathbb{R}^{N_{P}\times3}$ with $N_{P}$ the number of points, $M_{tp} \in (V, E)$ with $V \in \mathbb{R}^{N_{V}\times3}$ the 3D positions of vertices and $E \in \mathbb{Z}^{N_E\times3}$ the set of triangles. As shown in Figure \ref{fig:network_deform}, taking $P_{scn}$ and $V$ as inputs, our network regress the deformation vectors $D\in \mathbb{R}^{N_V\times3}$ of each vertex to deform $M_{tp}$ to the target mesh $M_{tgt}=(V', E)$, where $V'=V+D$. There are two challenges in this task. Firstly, the template mesh and observed point cloud are in different poses and scales, for which existing works \cite{wang20193dn,yifan2020neural} can not handle well by design. Secondly, it remains unknown how to fully leverage the structure and geometric information residing in the observed \textit{partial} point cloud to enhance the mesh deformation. To tackle the first challenge, we introduce a pose- and scale-invariant feature extraction backbone based on a 3DGC (3D graph convolution) network \cite{lin2020_3DGC}. We propose a novel multi-scale attention module for the second problem that can fully leverage the local and global geometric information residing in the partial point cloud to enhance mesh deformation.

\subsubsection{Pose- and Scale-Invariant Feature Extraction}
We utilize the shift- and scale-invariant 3DGC network as our feature extraction backbone to eliminate the pose and scale-invariant. 3DGC network is firstly applied to point cloud classification, object part segmentation \cite{lin2020_3DGC} and extended to visible point reconstruction in \cite{chen2021fsnet}. In this work, we further extend it to the mesh deformation problem. By designed, 3DGC is invariant in shift and scale (see \cite{lin2020_3DGC} for details). We utilize global max pooling to eliminate the influence of different orientations to obtain the final pose and scale-invariant global geometric features. Since the template and target mesh are in the normalized coordinate systems with the same pose and scale, we can easily leverage local and global geometric features from the template as geometric prior to deformation. As a guide to target shape deformation, geometric information from the observed scene point cloud is also required. One simple way is to add the scale- and pose- invariant global geometric feature from the scene point cloud as guidance. However, the global features lose local geometric details in the scene point cloud due to the max-pooling operation. While local geometric details are also helpful to enhance mesh deformation, we propose a novel multi-scale attentive geometry enhancement module to make full use of this information in the following section.

\subsubsection{Multi-Scale Geometry Enhancement with Transformers}
In the design of the encoder-decoder geometry feature extraction framework, local features are extracted with smaller perspective fields in the early stages. The perspective field grows larger when the network goes deeper and more global features are extracted. To guide the shape deformation better, both local and global geometric features from the observed target point cloud are crucial. Precisely, the early local features can guide the deformation of geometry details while the global ones guarantee global topology consistency. However, it is nontrivial to obtain corresponding local features from points in the scene to each vertex in template mesh, as they are not one-to-one aligned. To tackle the problem, we observe that corresponding parts of different object instances within the same category are in similar geometric structures, and the extracted geometric features are also similar. Therefore, we can utilize the similarity between local geometric features of each vertex in template mesh and each point in the scene to obtain the most related local features for deformation guidance. Such formulation is similar to the attention mechanism in Transformers \cite{attentionisallyouneed,wang2020linear_trans}, leading to a straightforward choice of using Transformer networks to serve our purpose. 

The Transformers first introduced in Natural Language Processing is good at capturing the long-term dependency, even on un-ordered sets. The key element in Transformers is the attention layer. Taking the input vectors, namely query $Q$, key $K$ and value $V$, the attention module retrieves information from $V$ according to the attention weight calculated from the dot product of $Q$ and $V$, i.e., $I = softmax(QK^T)V$ with $I$ the retrieved info. To reduce the quadratic computational cost of vanilla transformers \cite{attentionisallyouneed}, we apply the optimized Linear Transformer \cite{wang2020linear_trans} to extract the corresponding local geometric features from scene points to enhance features of the template mesh. Specifically, we take local features from the template mesh as $Q$ and local features from scene points as $K$ and $V$ and feed them to the Linear Transformers. The local geometrical features are then fed into the mesh deformation network header. We apply Linear Transformers on each encoding layer to retrieve multi-scale local to global geometric information from scene points. Another advantage of this mechanism is that the attention mechanism eases the negative effect of the partial-to-complete inconsistency. Specifically, invisible symmetric parts of the template mesh can also retrieve geometric information from the visible symmetric parts with the attention module to guide deformation.

\subsection{Differentiable Registration for Pose and Scale Estimation }
We have eliminated the shape inconsistency by our pose- and scale-invariant shape deformation network. In this section, we focus on the pose and scale estimation. Given the deformed mesh $M_{df}=(V^{'},E)$ in the normalized object coordinate space and the observed scene point cloud in the camera coordinate space, our task is to estimate a scale parameter, denoted $s$ to scale $M_{df}$ to its actual object size, and a rotation $R\in SO(3)$ as well as a translation $T\in SE(3)$ to transform $M_{df}$ to the camera coordinate system. After this step, the visible part of the scaled and transformed mesh $M_{df}^{cam}=(V^{cam}, E)$ in the camera coordinate should be aligned with scene points, where $V^{cam} = sRV^{'} + T$. Inspired by the success of differentiable registration in the point cloud alignment \cite{el2021unsupervisedr} field, we formulate the problem as point cloud registration but estimate an extra scale parameter. Specifically, we utilize differentiable sampling \cite{wang20193dn} to sample a point cloud from the surface of the deformed mesh and then align it to the scene point cloud by correspondence exploring and geometric fitting. The correspondence exploring establishes the 3D-3D correspondence between the two point clouds while the geometric fitting estimate pose and scale parameters by Umeyama algorithms \cite{umeyama1991least}. The pipeline is shown in the middle part of Figure \ref{fig:network_full}.

\subsubsection{Point-wise Feature Extraction \& Correspondence Exploring}
Before pose and scale estimation, we establish the correspondence between mesh points and scene points with geometric features. As shown in Figure \ref{fig:network_full}, we utilize two-point cloud networks, i.e. PointNet$++$ \cite{qi2017pointnet++} to extract point-wise features for the two point clouds respectively. Weights on the two PointNet$++$ are shared to be Siamese. Challenges of this part reside in the partial-to-complete and the pose and scale inconsistency between the two point clouds. Recently, PREDATOR \cite{huang2021predator} introduced a deep attention block to enable early information exchange on the final feature encoding layer. Specifically, Graph Neural Networks (GNN) \cite{wang2019DGNN} are first applied to aggregate contextual relations individually. Then a Transformer network is applied as a cross-attention module to enhance contextual information (see more details in \cite{huang2021predator}). We apply a similar technique but improve it in two ways. Firstly, GNN can only establish local attention with small perspective fields since only $K_{nn}$ nearest neighbors are linked to the graph. Therefore, we replace it with Linear Transformers that are good at capturing long-term dependency to establish global attention for better contextual aggregating. Secondly, in the cross attention part, we replace the vanilla transformers \cite{huang2021predator} with Linear Transformers to reduce the computational cost. 

With the extracted point-wise features, we can establish the correspondences between the two point clouds by finding points with closest features. For example, for a point $p$ with feature $f_p$ in the mesh point cloud, its corresponding point $p_q$ with feature $f_q$ on the scene point cloud can be found by $q_p = argmin_{q\in{P_{scn}}} D(f_p, f_q)$, where $D(p, q)$ is the function to calculate cosine distance. To eliminate false positives from the extracted correspondences, we follow \cite{el2021unsupervisedr} and utilize differentiable ratio test to recalculate the weights of each pairs (i.e., $w = 1 - \frac{D(p, q_{p,1})}{D(p, q_{p,2})}$) and select the top $K$ pairs for pose and size fitting. We also applied randomized subsets \cite{el2021unsupervisedr} techniques and split the $K$ pairs into $G$ correspondence groups, each consists of $K_g$ correspondence pairs: $\mathcal{C} = \{(p, q, w)_i: 1 \leq i \leq K_g\}$.
 
\subsubsection{Geometric Fitting for Pose and Size Estimation}
Given a set of correspondences $C=\{(p, q, w)_i: 1 \leq i \leq K_g\}$, we can estimate the pose and scale parameters with the Umeyama algorithm \cite{umeyama1991least}, which minimizes the squared loss:
\begin{equation}
    \label{eqn:umeyama}
    L_{Umeyama} = \sum_{i=1}^{K_g}||q_i - (sRp_i+T)||^2_2,
\end{equation}
where $s$ denotes the scale, $R$ the rotation and $T$ the translation. Among $G$ correspondence sets, we select the one that minimizes the loss as final prediction. 

\subsection{Differentiable Mesh Rendering}
In the first two steps, we have eliminated the shape, pose and scale inconsistency by differentiable deformation and registration. Here we utilize differentiable mesh rendering to eliminate the final partial-to-complete inconsistency. We first scale and transform the deformed normalized mesh from the object coordinate system to the camera coordinate system. We then utilize a differentiable mesh renderer \cite{ravi2020pytorch3d} to render the depth image. Finally, we lift the valid values in the depth image back to a point cloud to obtain the predicted visible object point cloud. In this way, we can enforce the geometric consistency between the predicted and the observed scene point cloud for supervision.

\subsection{Regularization}
We apply three loss functions to regularize the learning of shape deformation. 

\textbf{Deformed Shape Loss.} Following \cite{fan2017point,wang20193dn}, Chamfer distance is applied to regularize the shape deformation. One approach is to utilize point clouds sampled from the ground-truth mesh for supervision. While we want to eliminate the reliance on high-cost precise mesh models, we utilize coarse object point clouds for weak supervision. The coarse object point cloud is obtained from several real RGBD images of different views of the target object placed on marker boards, which are aggregated to be a complete object point cloud. It is further transformed to the same pose and normalized to the same scale as the template mesh (See the supplementary) for weak supervision. Given the coarse point cloud $P_{T}$ of the target object and the point cloud $P_{D}$ differentiably sampled \cite{wang20193dn} from the deformed mesh, the Chamfer distance can be calculated by
\begin{equation}
     \label{eqn:chamfer}
     \begin{split}
     L_{cd} = & \sum_{p_1 \in P_T}\min_{p_2 \in P_D}||p_1 - p_2||^2_2  +\sum_{p_2 \in P_D}\min_{p_1 \in P_T}||p_2 - p_1||^2_2 ,
     \end{split}
 \end{equation}
 
\textbf{Mesh Laplacian Loss.} Following 3DN \cite{wang20193dn}, we apply laplacian loss to restrict smooth deformation across the mesh surface and preserve geometric details: 
\begin{equation}
    \label{eqn:lap}
    L_{lpc} = \sum_i ||Lpc(M) - Lpc(M^*)||_2^2, 
\end{equation}
where $Lpc$ denotes mesh Laplacian operator, $M$ the original mesh, and $M^*$ the deformed one.

\textbf{Mesh Normal Consistency Loss.} 
To ensure the smoothness of the deformed mesh, a mesh normal consistency loss \cite{ravi2020pytorch3d} is also applied:
\begin{equation}
    \label{eqn:nrm}
    L_{nc} = \sum_{n_0 \in N} \sum_{n_1 \in N} \mathbb{I}(n_0, n_1)\cdot(1 - \
    cos{(n_0, n_1)}),
\end{equation}
where $N$ denotes normal vectors of mesh faces. $\mathbb{I}$ is an indication function equates to 1 when two normal vectors are from two neighboring faces, and 0 otherwise.

The total loss for shape deform is denoted as
\begin{equation}
    \label{eqn:deformloss}
    L_{df} = \lambda_{cd} L_{cd} + \lambda_{lpc} L_{lpc} + \lambda_{nc} L_{nc},
\end{equation}
where $\lambda_{*}$ denotes the weight of the corresponding loss.

We apply geometric consistency loss and correspondence loss to regularize the learning of the pose and scale prediction network. 

\textbf{Geometric Consistency Loss.} We utilize Chamfer distances to measure the geometric consistency between the predicted point cloud $P_{pred}$ and the observed scene point cloud $P_{scn}$. The calculation is similar to Equation \ref{eqn:chamfer} and we denote the geometric consistency loss as $L_{geo}$.

\textbf{Correspondence Loss.} Following \cite{el2021unsupervisedr}, we also apply weighted correspondence loss to enforce better correspondence exploring. Given the predicted correspondence sets $\mathcal{C} = \{(p, q, w)_i, 1 \leq i \leq K\}$ and the estimated scale and pose parameters ($s, R, T$), the weighted correspondence loss is defined as
 \begin{equation}
     \label{eqn:w_corres}
     \begin{split}
     L_{w\_corr} = \frac{1}{K}\sum_{i=1}^K w_i||q_i - (sRp_i+T)||_2^2.
     \end{split}
\end{equation}

We obtain the total loss functions for registration as follows:
\begin{equation}
    \label{eqn:poseloss}
    L_{pose} = \lambda_{geo} L_{geo} + \lambda_{w\_corr} L_{w\_corr},
\end{equation}
where $\lambda_*$ denotes the weight of each loss.

\section{Experiments}
\subsection{Benchmark Datasets}

\textbf{NOCS-REAL} \cite{wang2019normalized} is a real-world dataset collected for category-level object pose and size estimation. It contains six categories of hand-scale objects, with six unique instances each. Real-world images of 13 scenes are captured, with 4300 images of 7 scenes for training and 2750 images of 6 scenes for evaluation. The cluttered scenes and variety of object shapes make the dataset challenge.

\textbf{YCB-Video} \cite{calli2015ycb} is a popular benchmark dataset for instance-level object pose estimation. We select five typical objects from the dataset and split the training and testing set as in \cite{hodan2018bop}. 

\renewcommand{\arraystretch}{1.3}
\newcommand{\ycbC}{0.6}
\begin{table*}[tp]
    \centering
    \fontsize{7.}{7.}\selectfont
\begin{tabular}{L{1.4cm}ccccccccccc} 
\toprule
                                  & \multicolumn{1}{L{1.3cm}}{}    & \multicolumn{3}{c}{Training Data} & \multicolumn{7}{c}{mAP $\uparrow$}                                                  \\ \cmidrule(lr){3-5} \cmidrule(lr){6-12}
Learning Scheme                   & Method                   & real & syn. & type & $IoU_{25}$ & $IOU_{50}$ & $IOU_{75}$ & $5^\circ2cm$ & $5^\circ5cm$ & $10^\circ2cm$ & $10^\circ5cm$ \cr\hline
\multirow{8}{1cm}{Fully-Supervised} & SPGA \cite{Chen_2021_SGPA}  & $\checkmark$ & $\checkmark$         & R\&D   & -   & 80.1   & 61.9   & 35.9       & 39.6       & 61.3        & 70.7        \\
                                  & DualPoseNet \cite{Lin_2021_dualposenet} & $\checkmark$  & $\checkmark$  & R\&D   & -     & 79.8   & 62.2   & 29.3       & 35.9       & 50.0        & 66.8        \\
                                  & FS-Net \cite{chen2021fsnet} & $\checkmark$ &         & D  & -     & 92.2   & 63.5   & -          & 28.2       & -           & 64.6        \\
                                  & CASS \cite{chen2020CASS} & $\checkmark$ & $\checkmark$ & R\&D   & -     & 77.7   & -      & -          & 23.5       & -           & 58.0        \\
                                  & SPD \cite{tian2020SPD} & $\checkmark$ & $\checkmark$ & R\&D   & 83.0  & 77.3   & 53.2   & 19.3       & 21.4       & 43.2        & 54.1        \\
                                  & NOCS \cite{wang2019normalized} & $\checkmark$ & $\checkmark$  & R\&D   & 84.9  & 80.5   & -      & -          & 9.5        & -           & 26.7        \\
                                  & SPD* \cite{tian2020SPD} & $\checkmark$ &          & R\&D   & 82.8  & 63.3   & 39.2   & 3.6        & 4.8        & 19.0        & 26.0        \\
                                  & NOCS \cite{wang2019normalized} & $\checkmark$ &         & R\&D   & 61.9  & 47.5   & -      & -          & 6.5        & -           & 18.5        \\ \midrule
Traditional                            & ICP \cite{besl1992_ICP}       &          &          & D  & 60.2 & 43.8  &  6.3  & 0.2    &  1.8    &  0.7    &  2.1     \\ \midrule
\multirow{3}{1cm}{Self-Supervised}  & Ours-P              & $\checkmark$ &          & D  &76.1  &34.1  &8.1	&0.4	&1.8	&0.5	&11.9       \\
                                  & Ours-M               & $\checkmark$ &          & D  & 83.5  & 58.7   & 32.0   & 4.7        & 5.6        & 13.7        & 17.4        \\
                                  & Ours-M + ICP \cite{besl1992_ICP}  & $\checkmark$  &          & D  & 83.2  & 65.0   & 38.6   & 17.9       & 25.0       & 28.0        & 42.6           \\
                                  \bottomrule
\end{tabular}
    \caption{Quantitative results on the NOCS-REAL dataset. Our self-supervised depth-only framework outperforms traditional ICP and some fully-supervised algorithms. syn.: synthesis data; R\&D: RGBD; D: depth only; Ours-P: point-based; Ours-M: mesh-based; -: results not reported.  SPD*: We use the official code\protect\footnotemark of SPD and train on REAL275 for 50 epochs as ours. 
    }
    \label{tab:NOCS_real}
  \vspace{-0.8em}
\end{table*}

\footnotetext{https://github.com/mentian/object-deformnet}

\subsection{Evaluation Metrics}
For category-level pose and size estimation, we follow previous works \cite{wang2019normalized,chen2021fsnet} and adopt the average precision of Intersection-over-Union (IoU) and n$^\circ$m$cm$ as our evaluation metric. The $IoU_{x}$ metrics evaluate pose and size entangled 3D object detection performance, with $x\in \{25\%, 50\%, 75\%\}$ different acceptance IoU threshold. The pose-only $D^\circ Mcm$ metric evaluates the performance of 6D pose estimation. The prediction is considered accurate when the rotation error is less than $D^\circ$, and the translation error is less than $M$ center meter. The rotation error around the symmetry axis is ignored for symmetrical object categories. 
For instance-level 6D pose estimation, we follow \cite{xiang2017posecnn,he2020pvn3d,he2021ffb6d} and report the ADD-S AUC (area under the accuracy-threshold curve) metric for symmetrical objects and ADD AUC for non-symmetrical ones.

\subsection{Training and Implementation}
\textbf{Network architecture.} We apply the part segmentation version of the 3DGCN \cite{lin2020_3DGC} and PointNet$++$ \cite{qi2017pointnet++}. Each Linear Transformer \cite{wang2020linear_trans} has 4 attention heads.

\textbf{Training details.} We first train the mesh deformation network then fix the weight and train the registration network. We train each network with Adam \cite{kingma2014adam} optimizer by 50 epochs. The initial learning rate is $1e$-$4$ for deformation network and $2e$-$5$ for registration, and are halved every 10 epochs. We empirically set $\lambda_{cd}$, $\lambda_{lpc}$, $\lambda_{nc}$ to be $3$, $0.1$, $0.01$ in Formula \ref{eqn:deformloss} and $\lambda_{geo}, \lambda_{w_corr}$ to be $10$, $0.1$ in Formula \ref{eqn:poseloss}. For correspondence exploring, we set $K$ to be 400, and set $G$, $K_g$ to be 10 and 40 during training and 100 and 4 during testing.

\begin{figure}
  \centering
  \hspace{-3mm}\includegraphics[scale=0.16]{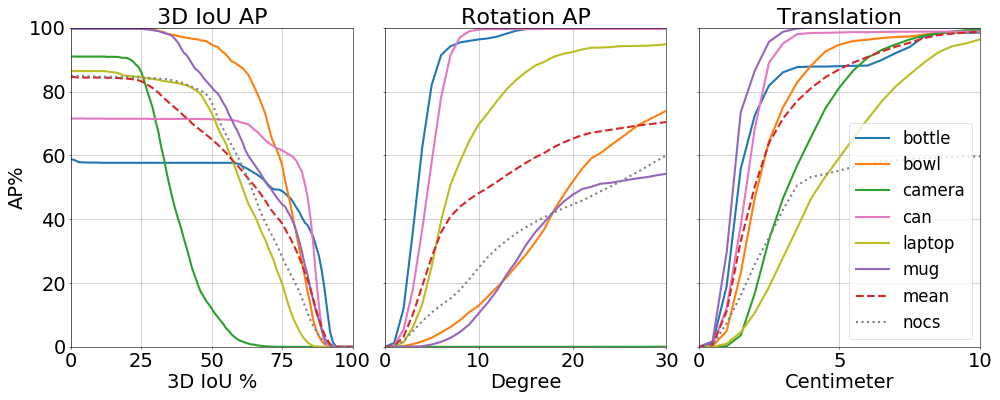}
  \caption{
    Results of our refined pipeline on the NOCS-REAL test set. The average precision of 3D IoU, rotation, and translation in different thresholds are reported.
  }
  \label{fig:mAP}
\end{figure}

\begin{figure}
\begin{minipage}{0.48\linewidth}
    \centering
    \includegraphics[scale=0.18]{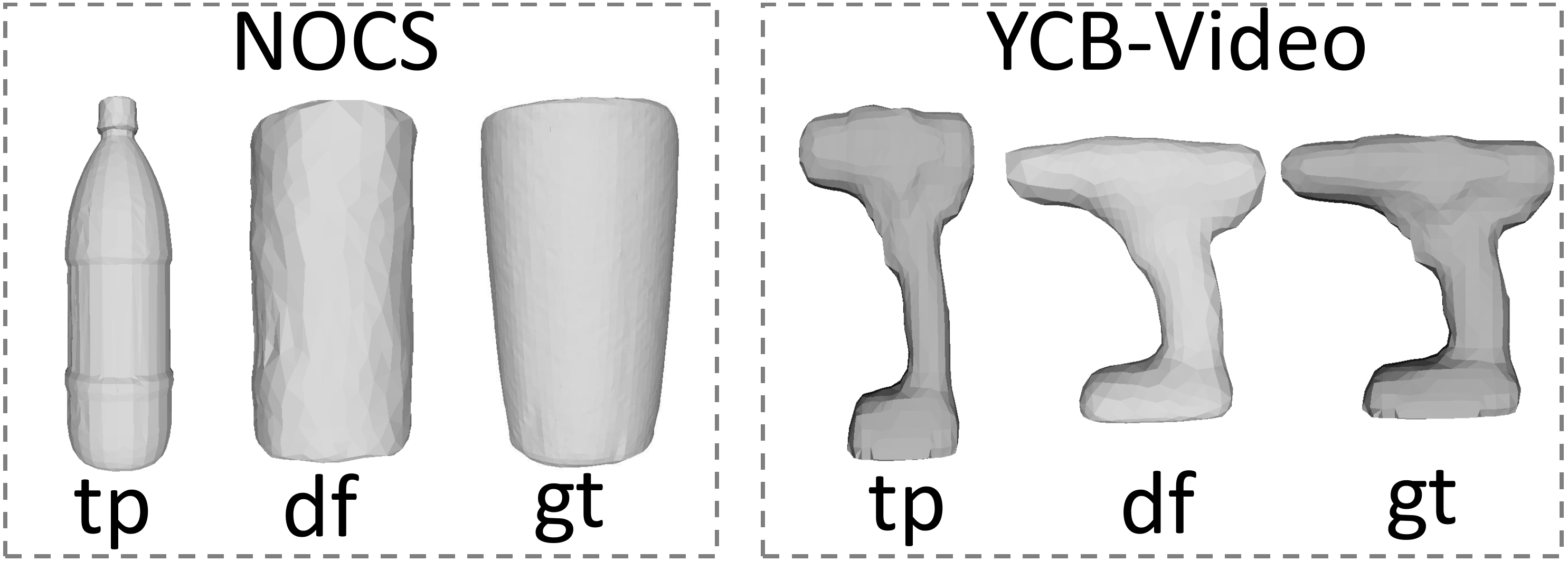}
    \caption{
        Qualitative results on mesh deformation. tp: category template mesh; df: deformed mesh by our network; gt: ground truth.
    }
    \label{fig:vis_df}
\end{minipage}
\hspace{3mm}
\begin{minipage}{0.48\linewidth}
    \centering
    \includegraphics[scale=0.46]{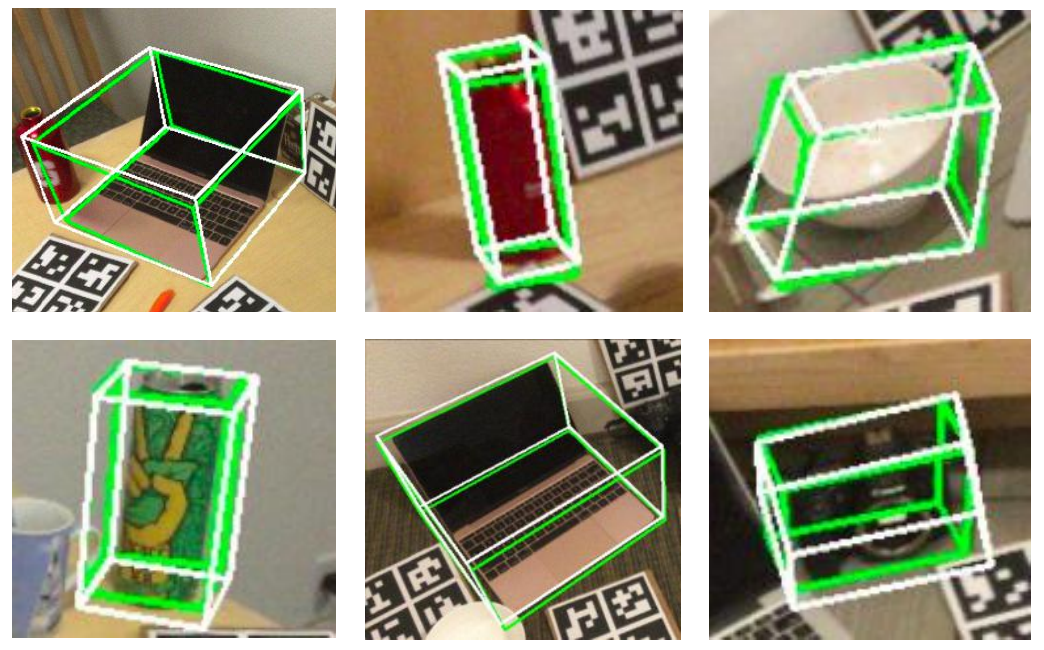}
      \caption{
        Qualitative results on NOCS-REAL. The predicted poses are in green, and the ground truth in white.
      }
     \label{fig:vis_pose}
\end{minipage}
\end{figure}

\subsection{Evaluation on Benchmark Datasets}

\textbf{Category-level pose and size estimation.} Table \ref{tab:NOCS_real} shows the quantitative evaluation results on the NOCS-REAL dataset. We compare our self-supervised framework with existing fully-supervised ones. Besides, we apply ICP on the template mesh as a simple baseline that does not require ground-truth pose labels.
As shown in the table, our mesh-based self-supervised approach (Ours-M) surpasses ICP by a large margin, i.e. +25.7\% on $IoU_{25}$ and +15.3\% on $10^{\circ}5cm$. We also obtained competitive or even better results against some supervised approaches that only train on real data as ours, i.e. +21.6\% on $IoU_{25}$ compared with NOCS \cite{wang2019normalized} and +0.8\% on $5^{\circ}5cm$ compared with SPD \cite{tian2020SPD}. Amazingly, since our full framework is designed to eliminate the entangled shape, pose, scale, and partial-to-complete inconsistencies between the category template shape and the scene point cloud, we find that applying simple ICP on our rendered depth and the scene point cloud for refinement further improve the accuracy by large margins, proving the efficacy of our network to eliminate these challenging inconsistencies. Moreover, our ICP refined version shows the possibility of eradicating labor-intensive pose labels for training as it already surpasses many recently proposed fully-supervised methods (i.e., NOCS-2019, SPD-2020, CASS-2020) on the strict $5^\circ5cm$ metric, even when they are trained with extra synthesis data that require precise CAD models for rendering. Detailed results reported following NOCS \cite{wang2019normalized} are in Figure \ref{fig:mAP}. Some qualitative results are shown in Figure \ref{fig:vis_pose}.

\textbf{Shape deformation.} For mesh deformation, we choose the mesh from ShapeNet \cite{chang2015shapenet} with smallest CD distance to the prior point shape in SPD \cite{tian2020SPD} as template mesh (TPM). As shown in Table \ref{tab:Shape Deformation}, our pose- and scale-invariant shape deformation network advances the 3DN \cite{wang20193dn} baseline by 29\% on shape deformation and also outperforms some point cloud based shape deformation baselines. 

\textbf{Shape deformation and pose estimation on the YCB-Video.} We first add noises to the provided CAD models by box-cage deformation in FS-Net \cite{chen2021fsnet} to be templates. We mask out object point clouds from the scene with ground-truth segmentation for training and testing. Results of our shape deformation and pose estimation are shown in Table \ref{tab:YCB_shape} \& \ref{tab:YCB_pose} respectively. Qualitative results are shown in Figure \ref{fig:vis_df} and the supplementary.

\newcommand{\OlC}{0.6}
\begin{table}[tp]
    \begin{minipage}{0.48\linewidth}
        \centering
        \fontsize{7.}{7.}\selectfont
        \begin{tabular}{lC{\OlC cm}C{\OlC cm}C{\OlC cm}C{\OlC cm}C{\OlC cm}C{\OlC cm} }
            \toprule
            Object ID &1	&2	&5	&10	 &15       & Mean          \\ \midrule
            Template & 5.91 & 6.31 & 3.15 & 3.31 & 4.73 & 4.68 \\
            Ours  & 0.18 & 0.18 & 0.12 & 0.09 & 0.13 & 0.14 \\
            \bottomrule
        \end{tabular}
        \caption{Quantitative results of shape deformation in the Chamfer Distance metric $\downarrow$ ($\times10^{-3}$) on YCB-Video.
        }
        \label{tab:YCB_shape}
    \end{minipage}
    \hspace{3mm}
    \begin{minipage}{0.48\linewidth}
        \centering
        \fontsize{7.}{7.}\selectfont
        \begin{tabular}{lC{\OlC cm}C{\OlC cm}C{\OlC cm}C{\OlC cm}C{\OlC cm}C{\OlC cm} }
            \toprule
            Object ID &1	&2	&5	&10	 &15       & Mean          \\ \midrule
            ICP  & 17.3 & 17.6 & 32.6 & 25.9 & 24.0 & 23.5 \\
            Ours & \textbf{74.8} & \textbf{66.5} & \textbf{83.0} & \textbf{80.9} & \textbf{68.2} & \textbf{74.7} \\
            \bottomrule
        \end{tabular}
        \caption{Quantitative results of 6D object pose on the deformed YCB-Video dataset (ADD-(S) AUC $\uparrow$\cite{he2021ffb6d,xiang2017posecnn}).
        }
        \label{tab:YCB_pose}
    \end{minipage}
\end{table}

\subsection{Ablation Study}
In this subsection, we present extensive ablation studies on our design choices.

\newcommand{\SDFC}{0.7}
\newcommand{\fdeC}{0.7}
\begin{table}[tp]
\begin{minipage}{0.48\linewidth}
    \centering
    \fontsize{7.}{7.}\selectfont
    \begin{tabular}{lC{\SDFC cm}C{\SDFC cm}C{\SDFC cm}C{\SDFC cm}C{\SDFC cm}C{\SDFC cm} }
        \toprule
       & SPD \cite{tian2020SPD}  & SGPA \cite{Chen_2021_SGPA} & TPM & 3DN \cite{wang20193dn} & Ours /MS       & Ours          \\ \midrule

shape  & PC   & PC            & Mesh & Mesh & Mesh     & Mesh          \\ \midrule
bottle & 4.34 & 2.93          & 6.21 & 3.90 & 2.41     & \textbf{2.21} \\
bowl   & 1.21 & 0.89          & 0.88 & 0.98 & 0.66     & \textbf{0.55} \\
camera & 8.30 & \textbf{5.51} & 9.95 & 7.30 & 6.24     & 5.81          \\
can    & 1.80 & 1.75          & 3.15 & 1.83 & 1.82     & \textbf{1.70} \\
laptop & 2.10 & 1.62          & 4.39 & 1.81 & 1.14     & \textbf{0.97} \\
mug    & 1.06 & 1.12          & 0.88 & 1.10 & 0.98     & \textbf{0.77} \\ \midrule
mean   & 2.99 & 2.44          & 4.24 & 2.82 & 2.21     & \textbf{2.00} \\
        \bottomrule
    \end{tabular}
    \caption{Quantitative results of shape deformation in the Chamfer Distance metric $\downarrow$ ($\times10^{-3}$). TPM: category template mesh to be deformed; /MS: without multi-scale geometry enhancement; PC: point cloud.}
    \label{tab:Shape Deformation}
\end{minipage}
\hspace{3mm}
\begin{minipage}{0.48\linewidth}
  \centering
  \fontsize{7.}{7.}\selectfont
  \begin{tabular}{lC{\fdeC cm}C{\fdeC cm}C{\fdeC cm}C{0.85 cm}C{0.88 cm} }
    \toprule
         & \multicolumn{3}{c}{Module}          & \multicolumn{2}{c}{mAP$\uparrow$}   \\ \cmidrule(lr){2-4} \cmidrule(lr){5-6}
Method   & DEF.      & REG.       & REN.       & $IoU_{50}$         & $5^\circ5cm$          \\ \midrule
Ours     &     & $\checkmark$ & $\checkmark$    & 45.3 & 2.6 \\
Ours     & $\checkmark$ &   & $\checkmark$   & 40.5 & 1.9 \\
Ours     & $\checkmark$ & $\checkmark$ &            & 54.3          & 4.8          \\
Ours     & $\checkmark$ & $\checkmark$ & $\checkmark$ & \textbf{58.7}          & \textbf{5.6}          \\ \midrule
ICP      &           &            &            & 43.8          & 1.8          \\
Ours+ICP & $\checkmark$ &            &            & 48.7          & 4.8          \\
Ours+ICP & $\checkmark$ & $\checkmark$ &            & 59.2          & 10.4         \\
Ours+ICP & $\checkmark$ & $\checkmark$ & $\checkmark$ & \textbf{65.0}          & \textbf{25.0}   \\
    \bottomrule
  \end{tabular}
  \caption{Effect of each component in our pipeline. They are designed to disentangle different inconsistencies and benefits not only our self-supervised pipeline, but also the traditional ICP algorithm. DEF.: Deformation; REG.: Registration; REN.: Rendering.
  }
  \label{tab:EffectSteps}
\end{minipage}

\end{table}

\textbf{Effect of the deformation, registration, and rendering module.} In Table \ref{tab:EffectSteps}, we ablate different modules to study their effect. As shown in the table, for our framework, the deformation step eliminates the shape inconsistency and contributes 13.4\% mAP improvement on $IoU_{50}$ and 3\% on $5^\circ5cm$. The registration step eliminates pose and scale inconsistencies and outperforms direct pose regression (+18.2\% on $IoU_{50}$, +3.7\% on $5^\circ5cm$). Finally, the rendering module further eases the partial to complete inconsistency and advances the performance (+4.4\% on $IoU_{50}$, +0.8\% on $5^\circ5cm$). Besides, we further utilize ICP to validate the efficacy of each module in our pipeline for inconsistency elimination. The performance of ICP depends much on the initialization, which means it is sensitive to the shape, pose, scale, and partial-to-complete inconsistency between the two input point clouds. As shown in the table, when our framework step by step eases the above inconsistencies, the performance of ICP also improves progressively, proving the efficacy of each component.

\textbf{Effect of multi-scale geometry enhancement.} In Table \ref{tab:Shape Deformation}, compared with the one without multi-scale geometry enhancement (Ours /MS in the table), our full shape deformation pipeline improves the result by 9.5\%.

\textbf{Effect of Linear Transformers for point cloud registration.} We compare our Linear Transformer version for self- and cross-attention to the original GNN and vanilla transformers version in PREDATOR \cite{huang2021predator}. Our version gets +$1.3\%$ improvement on $IoU_{75}$ AP$\uparrow$ for pose and scale estimation and consumes less GPU memory, showing the efficacy of the modification.

\textbf{Is the shape deformation pose- and scale-invariant?} Yes. 
We transform the partial scene point cloud to the same pose and scale as the template mesh as inputs and then train the mesh deformation network, getting 1.98 mean CD distance, vs. 2.00 from pose- and scale-inconsistent inputs in Table \ref{tab:Shape Deformation}.

\textbf{Point-based or mesh-based approaches?} Some works deform shape \cite{tian2020SPD,Chen_2021_SGPA} or render depth \cite{el2021unsupervisedr} from point cloud. We implement a point-based version for comparison. Specifically, we randomly sample 2048 points from the mesh as template shape, remove losses denoted in Formula \ref{eqn:lap} \& \ref{eqn:nrm}, use a point renderer, and keep the rest as our mesh-based approach. We compare point-based (Ours-P) and mesh-based (Ours-M) pipelines in Table \ref{tab:NOCS_real}, finding the sparse point cloud causes artifacts on the rendered depth and has a big performance drop.

\section{Discussion}

We propose a novel self-supervised framework for category-level pose and size estimation via differentiable shape deformation, registration, and rendering. Thanks to the elimination of the challenging shape, pose, scale, and partial-to-complete inconsistencies by our designed modules, our refined pipeline outperforms traditional algorithms and some of the fully-supervised ones, showing the possibility of eradicating the labor-intensive ground truth pose labels. There are also limitations in this work. Firstly, though the requirement of ground truth pose labels are eradicated, a coarse target object point cloud is required to guide our shape deformation network. Secondly, despite good results to show the possibility of a pose-label free learning algorithm, there is still a performance gap between self-supervised and fully-supervised algorithms. We expect more future research to bridge the gap.

\appendix
\setcounter{figure}{6}
\setcounter{table}{5}
\setcounter{equation}{7}

\section{Appendix}
\subsection{Overview}
In this supplementary, we introduce the implementation details of the coarse object point cloud on the NOCS-REAL dataset and the shape deformation on the YCB-Video dataset in Sec. \ref{sec:imple_details}. We present more results in Sec. \ref{sec:more_res}, including model running time and visualization of mesh deformation and pose estimation on the YCB-Video dataset. In Sec. \ref{sec:more_ablation}, we further ablate the effect of template mesh selection strategy, different supervision signals for mesh deformation, and the effect of normal loss.

\subsection{Implementation Details}
\label{sec:imple_details}

\subsubsection{How to obtain coarse object point cloud?}
For supervision of mesh deformation, one strategy is to utilize the ground-truth mesh model for full supervision, as is done in \cite{wang20193dn}. However, the precise object CAD models are expensive to obtain and limit the capability of networks to scale to new categories. Instead, we utilize coarse object point cloud recovered from real RGBD images of several views of the target object for mesh deformation supervision. Specifically, for each object, we select 16 RGBD views of it in the real-world training set using the farthest rotation sampling algorithm. We first initialize a set of selected views with a random view from the training dataset. Then, another object view with the farthest rotation distance from views in the selected set is added. We repeat this procedure until 16 views of the target object are obtained. The distance between two rotations are calculated by the Euclidean distance between two unit quaternions \cite{ravani1983motion,huynh2009metrics} as follows:
\begin{equation}
    D(q_1, q_2) = \min\{||q_1 - q_2||, ||q_1+q_2||\},
\end{equation}
where $||\cdot||$ denotes the Euclidean norm and $q_1, q_2$ the unit quaternions. This strategy can ensure that each part of the object is captured by at least one camera view and can be recovered from the RGBD images.

\begin{figure}
    \centering
    \includegraphics[width=0.7\linewidth]{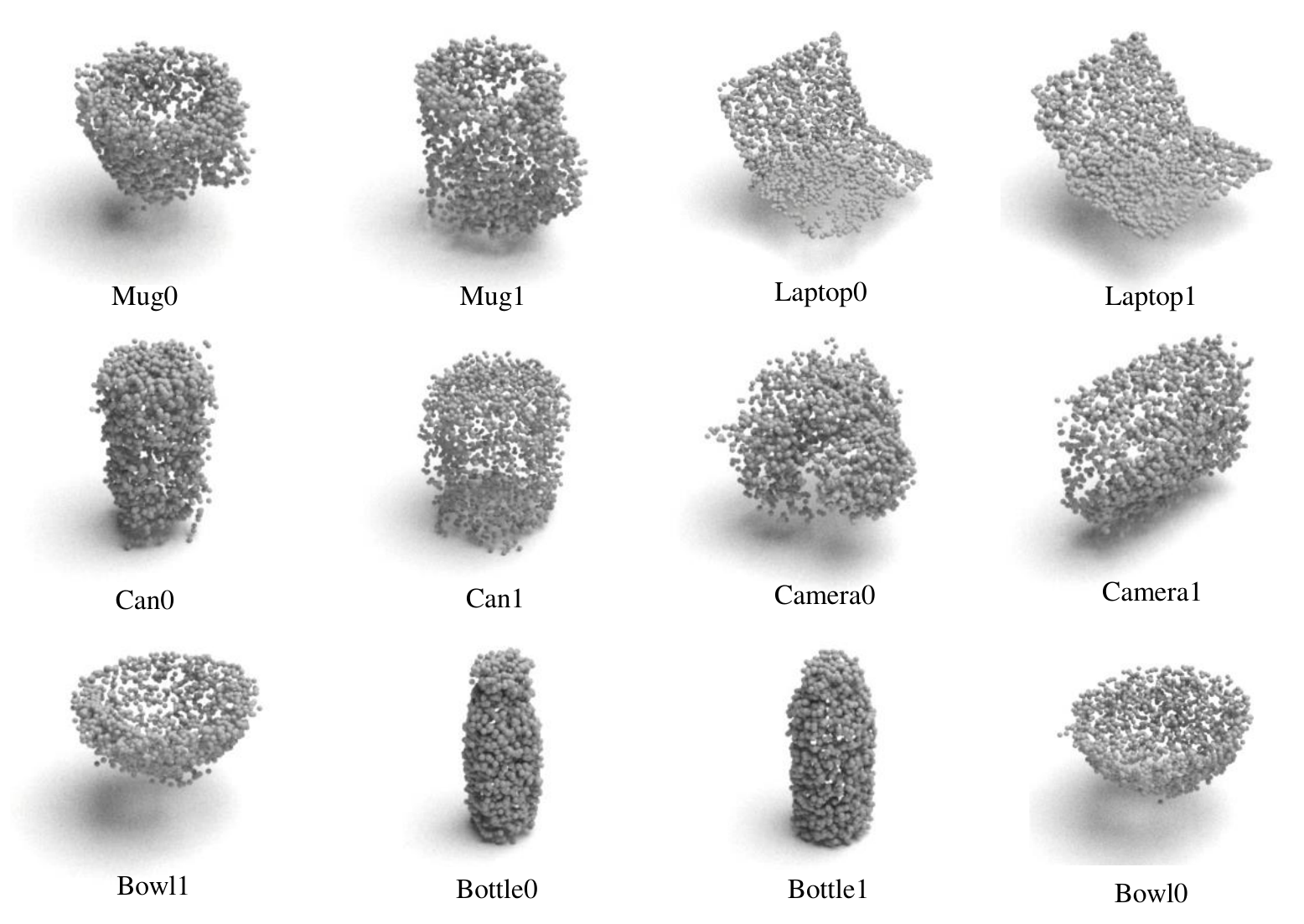}
    \caption{
        Visualization of some coarse object point clouds on the NOCS-REAL dataset.
    }
    \label{fig:coarse_pcld}
\end{figure}

For each selected view of the object, we crop out the object point cloud with the ground truth mask label and transform the object point cloud back to the normalized object coordinate system with the pose and scale parameters. We combine object point clouds from all the selected views to form a complete point cloud. In this way, the complete coarse object point cloud is in the same pose and scale as the template mesh. We also utilize MeanShift \cite{comaniciu2002mean} algorithm for removing the outliers caused by sensor noises. Visualization of some coarse object point clouds is shown in Fig. \ref{fig:coarse_pcld}.

For objects in a new category, we can quickly obtain the coarse object point clouds for weak supervision of mesh deformation with several views of each object similarly. These coarse object point clouds are easier to obtain than the precise CAD model as we do not require mesh reconstruction.

\subsubsection{Shape deformation in the YCB-Video dataset.}
To generate the category template meshes for shape deformation and pose and scale estimation on the YCB-Video dataset. We first add noise to the GT mesh model with box-cage deformation \cite{chen2021fsnet} algorithm. We then normalize the mesh to be in the normalized object coordinate space, where the diameter of each object is 1 meter. Visualization of the normalized template meshes and the ground-truth are shown in Fig. \ref{fig:vis_ycb_df}.

\subsubsection{Mesh Completion in the NOCS-Real dataset.}
We find that object meshes in NOCS-Real are not complete, with some faces missing. To better evaluate the accuracy of deformed complete mesh models in Table \ref{tab:Shape Deformation}, \ref{tab:random_nocs_template} and \ref{tab:tp_sup_sig}, we manually fill up those missing faces by Blender \footnote{https://www.blender.org/} mesh editing tools.

\subsection{More results}
\label{sec:more_res}

\subsubsection{Model Running Time}
We inference our model on a single NVIDIA RTX 2080Ti GPU. For each object, it takes an average time of 23ms for the mesh deformation network, 696ms for the point cloud registration network, and 8 ms for differentiable mesh rendering.

\subsubsection{Visualization of mesh deformation on the YCB-Video Dataset.}
\begin{figure}
\begin{minipage}{0.5\linewidth}
    \centering
    \includegraphics[width=1.\linewidth]{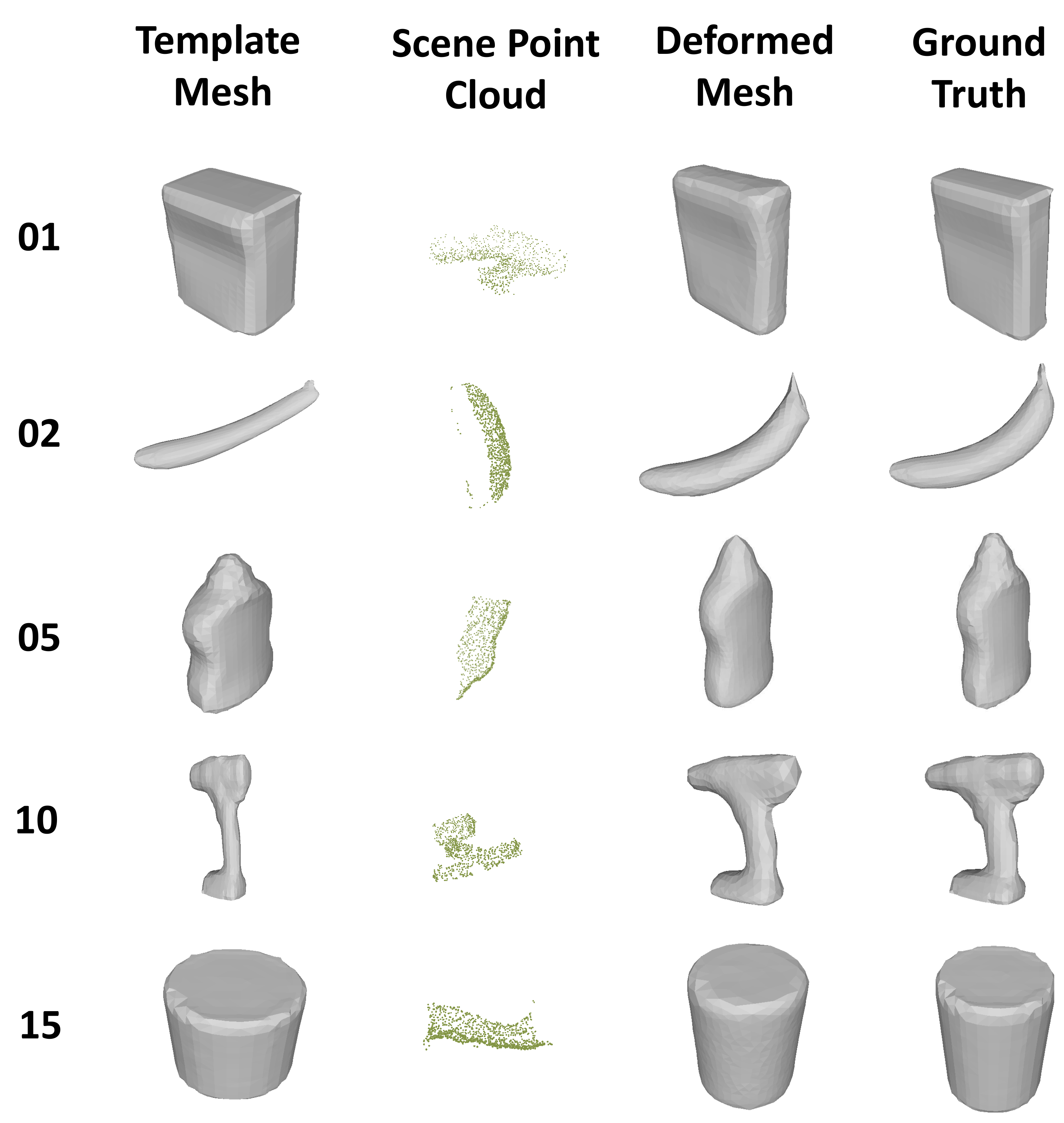}
    \caption{
        Visualization mesh deformation on the YCB-Video dataset.
    }
    \label{fig:vis_ycb_df}
    \vspace{-0.8em}
\end{minipage}
\begin{minipage}{0.5\linewidth}
    \centering
    \includegraphics[width=0.9\linewidth]{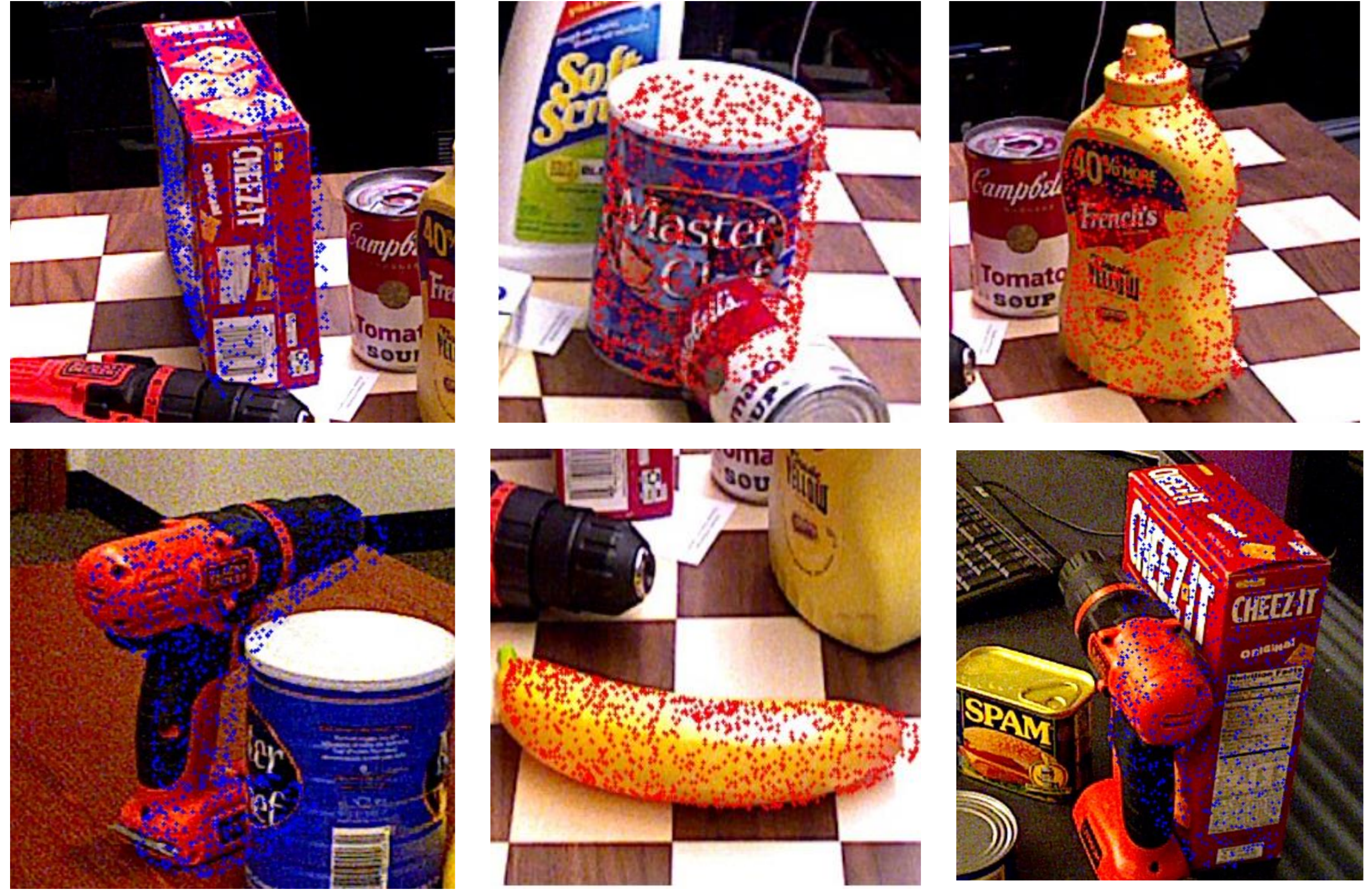}
    \caption{
        Visualization of pose and scale estimation on the YCB-Video dataset. We sample a point cloud from the surface of the normalized ground-truth model, scale and transform it to the camera coordinate system with predicted pose and scale parameters, and then project them on the scene images for visualization.
    }
    \label{fig:vis_ycb_pose}
\end{minipage}
\end{figure}

See Fig. \ref{fig:vis_ycb_df}. The category template meshes are generated from adding noises to the ground-truth mesh models by box-cage deformation algorithms \cite{chen2021fsnet}.

\subsubsection{Visualization of pose and scale estimation on the YCB-Video Dataset.} See Fig. \ref{fig:vis_ycb_pose}.

\subsection{More ablation study}
\label{sec:more_ablation}

\subsubsection{Effect of template mesh selection strategy.} To study the effect of different category template meshes for shape deformation, we randomly select one template mesh from ShapeNet for each class. As shown in Table \ref{tab:random_nocs_template}, our deformation network reduces the mean CD error of random template meshes by 50\%, which is comparable to the 53\% improvement on the prior template mesh in Table 4. It reveals the robustness of our mesh deformation network towards different template meshes.

\newcommand{\RNDDF}{0.9}
\renewcommand{\arraystretch}{1.4}
\begin{table}[]
    \centering
    \fontsize{8}{8}\selectfont
    \begin{tabular}{C{1.6 cm}C{\RNDDF cm}C{\RNDDF cm}C{\RNDDF cm}C{\RNDDF cm}C{\RNDDF cm}C{\RNDDF cm}C{\RNDDF cm}C{\RNDDF cm}}
    \toprule
Mesh & bot. & bow. & cam.  & can  & lap. & mug  & mean \\ \midrule
Template & 5.52 & 2.39 & 10.32 & 6.94 & 1.48 & 2.38 & 4.84 \\
Deformed & 2.63 & 0.64 & 6.54  & 2.69 & 0.69 & 1.22 & 2.40 \\
    \bottomrule
\end{tabular}
    \caption{Quantitative results of shape deformation in the Chamfer Distance metric $\downarrow$ ($\times10^{-3}$) with random selected category template mesh. }
    \label{tab:random_nocs_template}
    \vspace{-0.8em}
\end{table}

\renewcommand{\arraystretch}{1.4}
\begin{table}[]
    \centering
    \fontsize{8}{8}\selectfont
    \begin{tabular}{C{1.6 cm}C{\RNDDF cm}C{\RNDDF cm}C{\RNDDF cm}C{\RNDDF cm}C{\RNDDF cm}C{\RNDDF cm}C{\RNDDF cm}C{\RNDDF cm}C{\RNDDF cm}}
    \toprule
        & deform  & bot. & bow. & cam. & can  & lap. & mug  & mean          \\\midrule
Prior       & & 6.21 & 0.88 & 9.95 & 3.15 & 4.39 & 0.88 & 4.24          \\
Coarse PC   &\checkmark & 2.21 & 0.55 & 5.81 & 1.70 & 0.97 & 0.77 & 2.00 \\
GT Mesh     &\checkmark & 2.17 & 0.51 & 5.46 & 1.60 & 0.88 & 0.97 & 1.93 \\
    \bottomrule
\end{tabular}
    \caption{Quantitative results of shape deformation in the Chamfer Distance metric $\downarrow$ ($\times10^{-3}$) with different supervision signal. Prior: prior template mesh; Coarse PC: supervised by coarse object point cloud; GT Mesh: supervised by ground-truth mesh model.}
    \label{tab:tp_sup_sig}
\end{table}

\subsubsection{Effect of Different Supervision Signal on Mesh Deformation.} Since we want to relieve the reliance on precise mesh models that limit scaling up to new categories, we utilize the coarse object point cloud for mesh deformation supervision. We ablate the effect of shape supervision by ground truth mesh models and coarse object point cloud in Table \ref{tab:tp_sup_sig}.

\subsubsection{Effect of normal loss.} 
\begin{figure}
    \centering
    \includegraphics[width=0.36\linewidth]{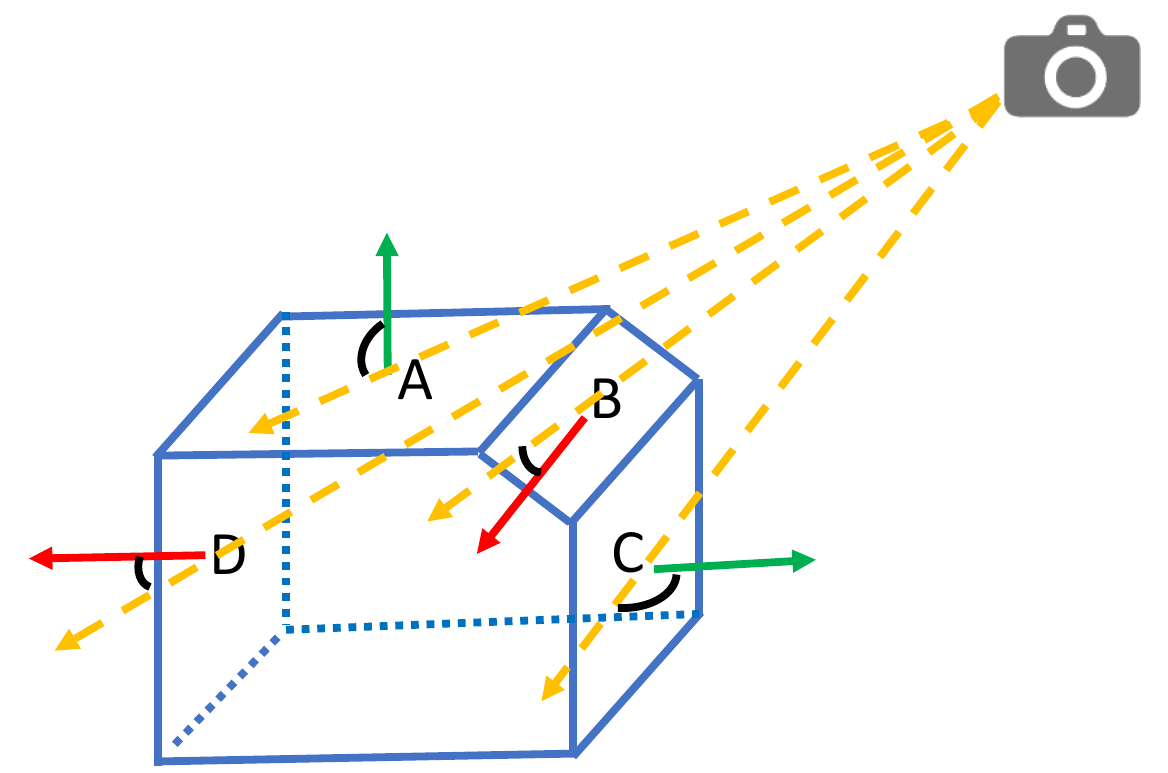}
    \caption{
       To speed up mesh rendering \cite{ravi2020pytorch3d}, faces with normal that are in the same direction as the camera viewing direction (i.e., the angle between two directions is less than 90 degrees) are regarded as invisible back faces (e.g., face D) and are not rendered. Without normal loss, some normal directions of neighboring faces may not be consistent, e.g., face B is different from A and C. It causes artifacts on rendered depth images, e.g., face B is not rendered due to the pruning.
    }
    \label{fig:nrm_ls}
\end{figure}

It contributes -$0.08$ mean CD error ($\times10^{-3}$)$\downarrow$ and $+0.5\%$ improvement on $IoU_{75}$ AP$\uparrow$. We also find that without the constrain of normal loss, the consistency of normal direction on the mesh surface is not guaranteed: some are pointing outside the object, and others are pointing inside the object. However, in the implementation of mesh rendering \cite{ravi2020pytorch3d}, the normal direction of each face is utilized to determine whether the faces are visible front faces or invisible back faces to speed up rendering, shown in Fig. \ref{fig:nrm_ls}. Some front faces with normal pointing inside the mesh model are pruned by mistake, causing artifacts in the rendered depth images. Since normals of all faces are pointing outside the object in the original template mesh, the constrain of normal loss during deformation keeps such tendency and relief the artifacts by large margins.

\clearpage
%
%
\bibliographystyle{splncs04}

\begin{thebibliography}{10}
\providecommand{\url}[1]{\texttt{#1}}
\providecommand{\urlprefix}{URL }
\providecommand{\doi}[1]{https://doi.org/#1}

\bibitem{besl1992_ICP}
Besl, P.J., McKay, N.D.: Method for registration of 3-d shapes. In: Sensor
  fusion IV: control paradigms and data structures. vol.~1611, pp. 586--606.
  International Society for Optics and Photonics (1992)

\bibitem{calli2015ycb}
Calli, B., Singh, A., Walsman, A., Srinivasa, S., Abbeel, P., Dollar, A.M.: The
  ycb object and model set: Towards common benchmarks for manipulation
  research. In: 2015 international conference on advanced robotics (ICAR). pp.
  510--517. IEEE (2015)

\bibitem{chang2015shapenet}
Chang, A.X., Funkhouser, T., Guibas, L., Hanrahan, P., Huang, Q., Li, Z.,
  Savarese, S., Savva, M., Song, S., Su, H., et~al.: Shapenet: An
  information-rich 3d model repository. arXiv preprint arXiv:1512.03012  (2015)

\bibitem{chen2020CASS}
Chen, D., Li, J., Wang, Z., Xu, K.: Learning canonical shape space for
  category-level 6d object pose and size estimation. In: Proceedings of the
  IEEE/CVF conference on computer vision and pattern recognition. pp.
  11973--11982 (2020)

\bibitem{Chen_2021_SGPA}
Chen, K., Dou, Q.: Sgpa: Structure-guided prior adaptation for category-level
  6d object pose estimation. In: Proceedings of the IEEE/CVF International
  Conference on Computer Vision (ICCV). pp. 2773--2782 (October 2021)

\bibitem{chen2021fsnet}
Chen, W., Jia, X., Chang, H.J., Duan, J., Shen, L., Leonardis, A.: Fs-net: Fast
  shape-based network for category-level 6d object pose estimation with
  decoupled rotation mechanism. In: Proceedings of the IEEE/CVF Conference on
  Computer Vision and Pattern Recognition. pp. 1581--1590 (2021)

\bibitem{chen2017multi}
Chen, X., Ma, H., Wan, J., Li, B., Xia, T.: Multi-view 3d object detection
  network for autonomous driving. In: Proceedings of the IEEE Conference on
  Computer Vision and Pattern Recognition. pp. 1907--1915 (2017)

\bibitem{collet2011moped}
Collet, A., Martinez, M., Srinivasa, S.S.: The moped framework: Object
  recognition and pose estimation for manipulation. The International Journal
  of Robotics Research  \textbf{30}(10),  1284--1306 (2011)

\bibitem{comaniciu2002mean}
Comaniciu, D., Meer, P.: Mean shift: A robust approach toward feature space
  analysis. IEEE Transactions on Pattern Analysis \& Machine Intelligence (5),
  603--619 (2002)

\bibitem{el2021unsupervisedr}
El~Banani, M., Gao, L., Johnson, J.: Unsupervisedr\&r: Unsupervised point cloud
  registration via differentiable rendering. In: Proceedings of the IEEE/CVF
  Conference on Computer Vision and Pattern Recognition. pp. 7129--7139 (2021)

\bibitem{fan2017point}
Fan, H., Su, H., Guibas, L.J.: A point set generation network for 3d object
  reconstruction from a single image. In: Proceedings of the IEEE conference on
  computer vision and pattern recognition. pp. 605--613 (2017)

\bibitem{geiger2012we}
Geiger, A., Lenz, P., Urtasun, R.: Are we ready for autonomous driving? the
  kitti vision benchmark suite. In: 2012 IEEE Conference on Computer Vision and
  Pattern Recognition. pp. 3354--3361. IEEE (2012)

\bibitem{he2021ffb6d}
He, Y., Huang, H., Fan, H., Chen, Q., Sun, J.: Ffb6d: A full flow bidirectional
  fusion network for 6d pose estimation. In: Proceedings of the IEEE/CVF
  Conference on Computer Vision and Pattern Recognition. pp. 3003--3013 (2021)

\bibitem{he2020pvn3d}
He, Y., Sun, W., Huang, H., Liu, J., Fan, H., Sun, J.: Pvn3d: A deep point-wise
  3d keypoints voting network for 6dof pose estimation. In: Proceedings of the
  IEEE/CVF Conference on Computer Vision and Pattern Recognition. pp.
  11632--11641 (2020)

\bibitem{hodan2017tless}
Hodan, T., Haluza, P., Obdr{\v{z}}{\'a}lek, {\v{S}}., Matas, J., Lourakis, M.,
  Zabulis, X.: T-less: An rgb-d dataset for 6d pose estimation of texture-less
  objects. In: 2017 IEEE Winter Conference on Applications of Computer Vision
  (WACV). pp. 880--888. IEEE (2017)

\bibitem{hodan2018bop}
Hodan, T., Michel, F., Brachmann, E., Kehl, W., GlentBuch, A., Kraft, D.,
  Drost, B., Vidal, J., Ihrke, S., Zabulis, X., et~al.: Bop: Benchmark for 6d
  object pose estimation. In: Proceedings of the European Conference on
  Computer Vision (ECCV). pp. 19--34 (2018)

\bibitem{huang2021predator}
Huang, S., Gojcic, Z., Usvyatsov, M., Wieser, A., Schindler, K.: Predator:
  Registration of 3d point clouds with low overlap. In: Proceedings of the
  IEEE/CVF Conference on Computer Vision and Pattern Recognition. pp.
  4267--4276 (2021)

\bibitem{huynh2009metrics}
Huynh, D.Q.: Metrics for 3d rotations: Comparison and analysis. Journal of
  Mathematical Imaging and Vision  \textbf{35}(2),  155--164 (2009)

\bibitem{jack2018learning}
Jack, D., Pontes, J.K., Sridharan, S., Fookes, C., Shirazi, S., Maire, F.,
  Eriksson, A.: Learning free-form deformations for 3d object reconstruction.
  In: Asian Conference on Computer Vision. pp. 317--333. Springer (2018)

\bibitem{jiang2020shapeflow}
Jiang, C., Huang, J., Tagliasacchi, A., Guibas, L., et~al.: Shapeflow:
  Learnable deformations among 3d shapes. arXiv preprint arXiv:2006.07982
  (2020)

\bibitem{kaskman2019homebreweddb}
Kaskman, R., Zakharov, S., Shugurov, I., Ilic, S.: Homebreweddb: Rgb-d dataset
  for 6d pose estimation of 3d objects. In: Proceedings of the IEEE/CVF
  International Conference on Computer Vision Workshops. pp.~0--0 (2019)

\bibitem{kingma2014adam}
Kingma, D.P., Ba, J.: Adam: A method for stochastic optimization. arXiv
  preprint arXiv:1412.6980  (2014)

\bibitem{krull2015learning}
Krull, A., Brachmann, E., Michel, F., Yang, M.Y., Gumhold, S., Rother, C.:
  Learning analysis-by-synthesis for 6d pose estimation in rgb-d images. In:
  Proceedings of the IEEE international conference on computer vision. pp.
  954--962 (2015)

\bibitem{kurenkov2018deformnet}
Kurenkov, A., Ji, J., Garg, A., Mehta, V., Gwak, J., Choy, C., Savarese, S.:
  Deformnet: Free-form deformation network for 3d shape reconstruction from a
  single image. In: 2018 IEEE Winter Conference on Applications of Computer
  Vision (WACV). pp. 858--866. IEEE (2018)

\bibitem{li2021leveraging}
Li, X., Weng, Y., Yi, L., Guibas, L., Abbott, A.L., Song, S., Wang, H.:
  Leveraging se (3) equivariance for self-supervised category-level object pose
  estimation. arXiv preprint arXiv:2111.00190  (2021)

\bibitem{Lin_2021_dualposenet}
Lin, J., Wei, Z., Li, Z., Xu, S., Jia, K., Li, Y.: Dualposenet: Category-level
  6d object pose and size estimation using dual pose network with refined
  learning of pose consistency. In: Proceedings of the IEEE/CVF International
  Conference on Computer Vision (ICCV). pp. 3560--3569 (October 2021)

\bibitem{lin2020_3DGC}
Lin, Z.H., Huang, S.Y., Wang, Y.C.F.: Convolution in the cloud: Learning
  deformable kernels in 3d graph convolution networks for point cloud analysis.
  In: Proceedings of the IEEE/CVF Conference on Computer Vision and Pattern
  Recognition. pp. 1800--1809 (2020)

\bibitem{marchand2015pose}
Marchand, E., Uchiyama, H., Spindler, F.: Pose estimation for augmented
  reality: a hands-on survey. IEEE transactions on visualization and computer
  graphics  \textbf{22}(12),  2633--2651 (2015)

\bibitem{qi2017pointnet++}
Qi, C.R., Yi, L., Su, H., Guibas, L.J.: Pointnet++: Deep hierarchical feature
  learning on point sets in a metric space. In: Advances in neural information
  processing systems. pp. 5099--5108 (2017)

\bibitem{ravani1983motion}
Ravani, B., Roth, B.: Motion synthesis using kinematic mappings  (1983)

\bibitem{ravi2020pytorch3d}
Ravi, N., Reizenstein, J., Novotny, D., Gordon, T., Lo, W.Y., Johnson, J.,
  Gkioxari, G.: Accelerating 3d deep learning with pytorch3d. arXiv preprint
  arXiv:2007.08501  (2020)

\bibitem{sundermeyer2020augmented}
Sundermeyer, M., Marton, Z.C., Durner, M., Triebel, R.: Augmented autoencoders:
  Implicit 3d orientation learning for 6d object detection. International
  Journal of Computer Vision  \textbf{128}(3),  714--729 (2020)

\bibitem{tian2020SPD}
Tian, M., Ang, M.H., Lee, G.H.: Shape prior deformation for categorical 6d
  object pose and size estimation. In: European Conference on Computer Vision.
  pp. 530--546. Springer (2020)

\bibitem{tremblay2018deep}
Tremblay, J., To, T., Sundaralingam, B., Xiang, Y., Fox, D., Birchfield, S.:
  Deep object pose estimation for semantic robotic grasping of household
  objects. arXiv preprint arXiv:1809.10790  (2018)

\bibitem{umeyama1991least}
Umeyama, S.: Least-squares estimation of transformation parameters between two
  point patterns. IEEE Transactions on Pattern Analysis \& Machine Intelligence
   \textbf{13}(04),  376--380 (1991)

\bibitem{attentionisallyouneed}
Vaswani, A., Shazeer, N., Parmar, N., Uszkoreit, J., Jones, L., Gomez, A.N.,
  Kaiser, {\L}., Polosukhin, I.: Attention is all you need. In: Advances in
  neural information processing systems. pp. 5998--6008 (2017)

\bibitem{wang2020self6d}
Wang, G., Manhardt, F., Shao, J., Ji, X., Navab, N., Tombari, F.: Self6d:
  Self-supervised monocular 6d object pose estimation. In: European Conference
  on Computer Vision. pp. 108--125. Springer (2020)

\bibitem{wang2019normalized}
Wang, H., Sridhar, S., Huang, J., Valentin, J., Song, S., Guibas, L.J.:
  Normalized object coordinate space for category-level 6d object pose and size
  estimation. In: Proceedings of the IEEE Conference on Computer Vision and
  Pattern Recognition. pp. 2642--2651 (2019)

\bibitem{wang2021CRNet}
Wang, J., Chen, K., Dou, Q.: Category-level 6d object pose estimation via
  cascaded relation and recurrent reconstruction networks. arXiv preprint
  arXiv:2108.08755  (2021)

\bibitem{wang2020linear_trans}
Wang, S., Li, B.Z., Khabsa, M., Fang, H., Ma, H.: Linformer: Self-attention
  with linear complexity. arXiv preprint arXiv:2006.04768  (2020)

\bibitem{wang20193dn}
Wang, W., Ceylan, D., Mech, R., Neumann, U.: 3dn: 3d deformation network. In:
  Proceedings of the IEEE/CVF Conference on Computer Vision and Pattern
  Recognition. pp. 1038--1046 (2019)

\bibitem{wang2019DGNN}
Wang, Y., Sun, Y., Liu, Z., Sarma, S.E., Bronstein, M.M., Solomon, J.M.:
  Dynamic graph cnn for learning on point clouds. Acm Transactions On Graphics
  (tog)  \textbf{38}(5),  1--12 (2019)

\bibitem{xiang2017posecnn}
Xiang, Y., Schmidt, T., Narayanan, V., Fox, D.: Posecnn: A convolutional neural
  network for 6d object pose estimation in cluttered scenes. arXiv preprint
  arXiv:1711.00199  (2017)

\bibitem{xu2018pointfusion}
Xu, D., Anguelov, D., Jain, A.: Pointfusion: Deep sensor fusion for 3d bounding
  box estimation. In: Proceedings of the IEEE Conference on Computer Vision and
  Pattern Recognition. pp. 244--253 (2018)

\bibitem{yifan2020neural}
Yifan, W., Aigerman, N., Kim, V.G., Chaudhuri, S., Sorkine-Hornung, O.: Neural
  cages for detail-preserving 3d deformations. In: Proceedings of the IEEE/CVF
  Conference on Computer Vision and Pattern Recognition. pp. 75--83 (2020)

\bibitem{yumer2016learning}
Yumer, M.E., Mitra, N.J.: Learning semantic deformation flows with 3d
  convolutional networks. In: European Conference on Computer Vision. pp.
  294--311. Springer (2016)

\end{thebibliography}

\end{document}